\pdfoutput=1

\documentclass[11pt]{article}

\usepackage{acl}

\usepackage{times}
\usepackage{latexsym}

\usepackage[T1]{fontenc}

\usepackage[utf8]{inputenc}

\usepackage{microtype}

\usepackage{inconsolata}

\usepackage{graphicx}

\usepackage{booktabs}
\usepackage{multirow}
\usepackage{pifont}
\usepackage{soul}
\usepackage{array}
\usepackage{amsmath}

\title{Towards Goal-oriented Prompt Engineering \\ for Large Language Models: A Survey}

\author{{\bf Haochen Li}\ \ \ \ 
    {\bf Jonathan Leung}\ \ \ \ 
    {\bf Zhiqi Shen}  \\
     Nanyang Technological University, Singapore \\
     \texttt{\{haochen003, jonathan008, zqshen\}@ntu.edu.sg
     }}

\begin{document}
\maketitle
\begin{abstract}

Large Language Models (LLMs) have shown prominent performance in various downstream tasks and prompt engineering plays a pivotal role in optimizing LLMs' performance. 
This paper, not only as an overview of current prompt engineering methods, but also aims to highlight the limitation of designing prompts based on an anthropomorphic assumption that expects LLMs to think like humans. 
From our review of 50 representative studies, we demonstrate that a goal-oriented prompt formulation, which guides LLMs to follow established human logical thinking, significantly improves the performance of LLMs. 
Furthermore, We introduce a novel taxonomy that categorizes goal-oriented prompting methods into five interconnected stages and we demonstrate the broad applicability of our framework. 
With four future directions proposed, we hope to further emphasize the power and potential of goal-oriented prompt engineering in all fields. 

\end{abstract}

\section{Introduction}

Large Language Models (LLMs) have garnered significant interest with their ability to assimilate vast amounts of information from large corpora of data. They have demonstrated proficiency in open-domain dialogue~\cite{proactive-cot,gdp-zero}, reasoning~\cite{cot,tot,got}, planning~\cite{llm-planner,llm+p}, and many more tasks. The predominant approach to leveraging LLMs for downstream tasks involves crafting tailored text prompts to tap into LLMs' potential~\cite{promptingframeworksurvey}. However, the effectiveness of LLMs is contingent upon the prompting strategy employed, leading to performance disparities~\cite{zhao2021calibrate}. Such a phenomenon has given rise to the field of prompt engineering, which investigates the optimal formulation of prompts for specific tasks.

To prompt LLMs appropriately, it is important to first understand the deviation in the human way of thinking and LLMs' way of processing information. Humans are taught by experience to approach complex objectives by breaking down the main goal into more manageable sub-goals, a strategy supported by goal-setting theory~\cite{austin1996goal}. In interacting with LLMs, which often display human-like conversational abilities, we might be tempted to attribute human-like thought processes to them, expecting them to know from the start that they should decompose complex problems into simpler tasks. This anthropomorphic assumption does not hold for LLMs and can lead to unsatisfactory outcomes~\cite{abercrombie-etal-2023-mirages}. On the other hand, as shown in our literature review of 50 representative studies, if we design the prompts from a goal-oriented perspective that guides LLMs to mimic human thinking, LLMs' performance improves significantly. In the meantime, aligning LLM's behavior with human logic allows more human-interpretable answers from LLMs and results in more effective and reliable human-computer interaction.

\begin{figure*}[t]
\centering
\includegraphics[width=0.99\textwidth]{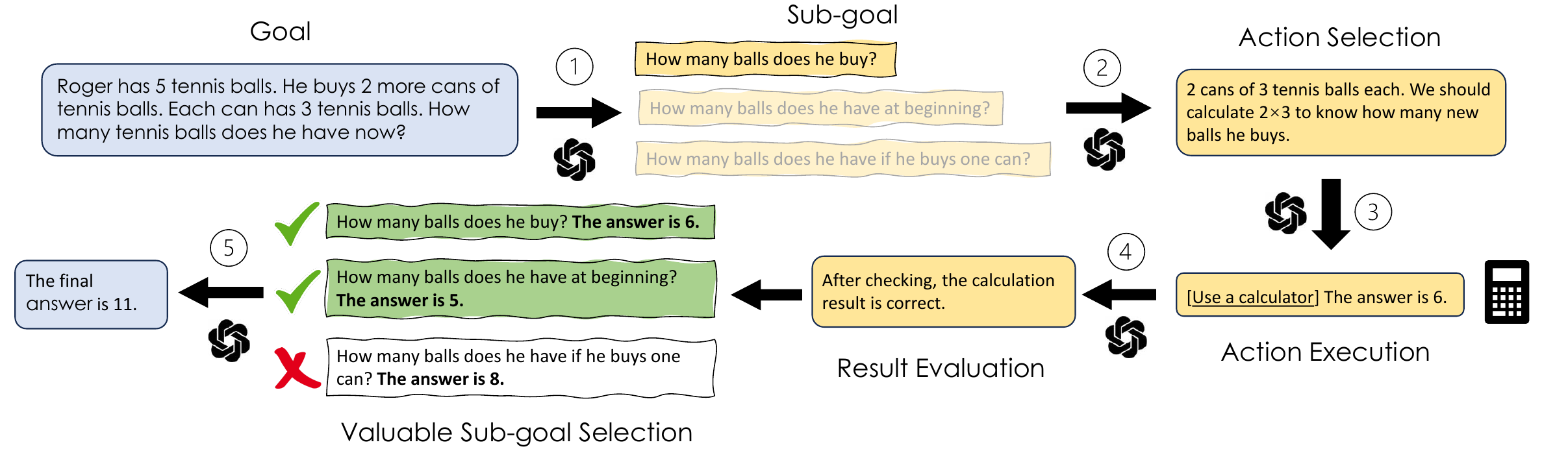}
\caption{An overview of the goal-oriented framework for prompting LLMs taking solving a math word problem as an example. (1) Decomposing \textbf{goal} into \textbf{sub-goal} sequences. (2) \textbf{Action} selection for attaining \textbf{sub-goals}. (3) Executing \textbf{action}s to get \textbf{sub-goal result}s. (4) Evaluating \textbf{sub-goal result}s. (5) Further selection of valuable \textbf{sub-goals}. Note that stages (2)(3)(4) are taken for all the decomposed sub-goals.}
\label{overview}
\end{figure*}

While several surveys have explored LLM prompting, none have examined it from a goal-oriented perspective or explicitly summarized the task-agnostic internal steps involved in designing and refining prompting procedures\footnote{We discuss related surveys in Appendix~\ref{relatedsurvey}.}.
In this paper, we propose the first goal-oriented taxonomy of prompting methods, which classifies goal-oriented prompting methods into goal decomposition, action selection, action execution, sub-goal result evaluation, and sub-goal selection, as shown in Fig.~\ref{overview}. We further classify the surveyed methods by their targeted task and analyze their respective performance to provide readers with a better estimate of the current progress of goal-oriented prompt engineering. Finally, we offer four promising future directions, including hierarchical decomposition, efficiency problems, the synergy of stages, and broader applications.

\section{Preliminary}

In this section, we provide a definition of goal-oriented prompt engineering. Before define goal-oriented prompting, we can first define standard prompting, the process of prompting LLM $Pr_{LLM}$ for a prediction on a given goal as:

\begin{equation}
    Pr(\mathcal{P}|\mathcal{G}) = \prod_{i=1}^{|\mathcal{P}|}Pr_{LLM}(p_i|\mathcal{G}, p_{<i})
\label{eq1}
\end{equation}

By introducing additional variables sub-goal $g$, action $a$, and sub-goal result $r$, we formally define goal-oriented prompt engineering as the application of a multi-stage strategy to guide LLMs in problem-solving by adhering to established human logic, as shown in Fig.~\ref{overview}.

In goal-oriented prompt engineering, the overall goal $\mathcal{G}$ is first decomposed into sub-goals $\{g_1,g_2,...,g_n\}$ (Fig.1(1)), then actions $\{a_1,a_2,...,a_n\}$ are selected to achieve corresponding sub-goals (Fig.1(2)). The final prediction is based on valuable sub-goal results $\{r_1,r_2,...,r_n\}$ (Fig.1(5)). The process could be described as:

\begin{align}
\small
   Pr(\mathcal{P}|\mathcal{G}) &= \sum_{i=1}^nPr(P|r_i,a_i,g_i,\mathcal{G}) \\
   & Pr(r_i|a_i,g_i,\mathcal{G})Pr(a_i|g_i,\mathcal{G})Pr(g_i|\mathcal{G}) \nonumber
\end{align}

If LLMs are applied for the implementation of a stage, we could follow Eq.~\ref{eq1} to get the output. We additionally introduce the action execution (Fig.1(3)) and sub-goal evaluation stages (Fig.1(4)) to ensure the correctness of sub-goal results.

\section{Taxonomy of Methods}

In this section, we go into detail about the five steps described in our goal-oriented framework in Fig.~\ref{overview}. 

\subsection{Decomposing Goal into Sub-goal Sequences}
\label{sec:stage1}
Decomposing a high-level goal into sub-goals is particularly useful for complex problems where a straightforward answer isn't sufficient. In this section, we introduce the methods to decompose goals and improve the performance of decomposition.
\paragraph{Iterative decomposition.} 
Iterative decomposition generates a sub-goal, gets sub-goal results, and repeats this process with the knowledge of the previous sub-goal and sub-goal results. 
Chain-of-thought prompting (CoT)~\cite{cot} can be considered the first work that follows iterative decomposition. With either ``\textit{Let's think step by step.}''~\cite{zero-shot-cot} in the prompt, or intermediate reasoning steps in in-context examples, LLMs can imitate the decomposition process and solve the problem step by step. Here, the intermediate reasoning steps can be considered as sub-goals, and they are sequentially connected to form a sub-goal sequence. CoT implicitly follows iterative decomposition as LLMs generate tokens in an autoregressive way, with the next sub-goal decided based on previous content. 

There are also works that explicitly ask LLMs to follow the iterative decomposition. DecomP~\cite{decomposed-prompting}, Successive Prompting~\cite{successive-prompt}, and Recursion-of-Thought (RoT)~\cite{rot} are three works that repeatedly prompt LLMs to ask follow-up sub-questions for background knowledge. 
Self-ask~\cite{self-ask} improves efficiency by designing a template for follow-up questions in advance.
With that template, LLMs can generate all essential questions and their answers by prompting once. Besides, in this paper, the authors empirically show that LLMs are often wrong when asking them to answer a complex question even if they know the true answer of all needed sub-questions. This finding indicates the significance of decomposing complex goals into simple sub-goals for LLMs. 

\paragraph{Plan-then-execute decomposition.}
In contrast to iterative decomposition, plan-then-execute decomposition methods generate the sub-goal sequence all at once, which means that the latter sub-goal will not be affected by the result of the former ones.
For example, Least-to-most prompting~\cite{least-to-most} only prompts LLMs two times, one for generating a plan to decompose the goal into a sub-goal sequence, and the other one for executing the plan. Plan-and-solve prompting~\cite{plan-and-solve} further improves the efficiency of Least-to-most prompting by merging the generation of the plan and execution of the plan into one output. DEPS~\cite{deps} and GITM~\cite{gitm} are decomposition methods designed for Minecraft, an open-world game in which an agent can craft different items and tools. In this game, obtaining base materials that are needed to craft a target item can be viewed as sub-goals. While DEPS generates a plan to obtain required objects in sequence solely based on LLMs, GITM leverages a pre-defined sub-goal tree to help LLMs locate prerequisites more precisely.

\paragraph{External decomposition.}
The above two categories rely on LLM's knowledge to decompose the goal into sub-goals. However, due to the hallucination problem~\cite{hallucination}, LLMs sometimes generate seemingly plausible sub-goals but not grounded in reality. To ensure the accuracy of decomposition, LLM+P~\cite{llm+p} and SayPlan~\cite{sayplan} take advantage of classical planners. They use LLMs to translate goals written in natural language to planning domain definition language so that classical planners can deal with them. The output of planners will then be translated back into natural language by LLMs for execution.

\subsection{Action Selection for Attaining Sub-goals}
With sub-goals defined, another important step is action selection, choosing effective and valid actions to reach desired outcomes. 
CoT is a naive way for action selection where it is completely decided by LLMs themselves. However, due to the hallucination problem, 
actions from LLMs are often invalid. In this section, 
we classify the advanced action selection methods into two classes, constrain-then-select and select-then-mapping.

\paragraph{Constrain-then-select.}
Constrain-then-select method predefines an action space and then has LLMs select the actual action among the space. 
As in MWP~\cite{math-word-problem}, to deal with math word problems, the authors first employ an operation prediction module to predict the needed calculation operation (e.g. multiplication, summation). LLMs are then used to select the appropriate operands and complete the calculation.
PEARL~\cite{pearl} utilizes constrained action space like ``\textit{Finding the definition of A}'', ``\textit{Compare A and B}'', and ``\textit{Summarize A}'' for question answering over long documents.
DecomP~\cite{decomposed-prompting} selects actions from a set like ``\textit{split}'' and ``\textit{merge}'' for k-th letter concatenation, a symbolic reasoning task. In robot planning, given a task instruction, SayPlan~\cite{sayplan} applies semantic search to first identify a task-relevant subgraph from the whole 3D scene graph, which makes it easier for LLMs to plan based on the subgraph.
In dialogue systems, RLP~\cite{rlp}, SAFARI~\cite{safari}, and Cue-CoT~\cite{cue-cot} set the mental states of the LLM and the user, respectively, such as beliefs or desires, to guide action selection. Such a setting improves the context richness, coherence, and interactivity of LLM-based conversational systems. Similar to PEARL, ProCoT~\cite{proactive-cot} forms pre-defined action sets covering query clarification, topic transition, and negotiation strategy for dialogue systems. 

\paragraph{Select-then-mapping.}
Different from constrain-then-select, select-then-mapping first uses LLMs to generate actions solely based on their knowledge and then maps the generated action to the most similar one in the valid action space. Zero-shot planners~\cite{zero-shot-planner}, SALP~\cite{generating-executable-action}, and Re-Prompting~\cite{re-prompting} aim to solve agent planning problems in an interactive environment. In such virtual environments, there are only a few admissible actions that can be applied. However, the actions produced naively by LLMs often cannot be mapped exactly to those executable actions. To remedy this, researchers employ a text-embedding language model to translate LLM-generated actions into the most similar admissible actions by calculating cosine similarity.

\subsection{Executing Actions to Get Sub-goal Results}
\label{sec:action execution}

In CoT, LLMs rely on their knowledge for action execution to get the sub-goal result. Limited by hallucination, the results are sometimes wrong. For example, even if LLMs correctly select multiplication as the action to solve a math problem, the calculation result can be wrong. As~\cite{decomposed-prompting} claimed, some tasks may not be feasible to solve using only LLMs.
With such consideration, some works leverage external tools to guarantee a precise sub-goal result. 

Some studies specifically study one task~\cite{decomposed-prompting,pot,recmind}.
To answer open-domain questions, DecomP~\cite{decomposed-prompting} applies an ElasticSearch-based retrieval system to retrieve knowledge from certain knowledge bases like Wikipedia. Program-of-Thoughts (PoT)~\cite{pot}, Faithful CoT~\cite{faithfulcot}, PAL~\cite{pal}, MathPrompter~\cite{mathprompter}, LINC~\cite{linc}, and Logic-LM~\cite{logic-lm} translate reasoning processes into executable codes and then run them in an interpreter. In recommender systems, Recmind~\cite{recmind} and InteRecAgent~\cite{InteRecAgent} use SQL to search for the history of a user in the database while DOKE~\cite{doke} achieves it through knowledge graphs. 

There are additional studies that aim for multi-tasking.
Toolformer~\cite{toolformer} specifically studied which APIs to call and when to call them when using external tools. In their paper, several tools are adopted, including a calculator, a calendar, a BM25-based Wikipedia search engine, and a fine-tuned language model (LM). The authors pointed out that small LMs finetuned for specific tasks can outperform some LLMs on certain task. Following the idea of Toolformer, HuggingGPT~\cite{hugginggpt} further extends the available LMs to the scale of all LMs on open-sourced machine learning communities like HuggingFace. Specifically, in their paper, they evaluate HuggingGPT on nine types of NLP tasks, nine types of CV tasks, four types of audio tasks, and two types of video tasks. With the development of open-sourced communities, this work can potentially achieve all the sub-goals that can be described in language.

\subsection{Evaluating Sub-goal Results}

\begin{table}[t]
\centering
\small
\begin{tabular}{@{}lp{.65\columnwidth}@{}}
\toprule
Feedback Source                 & Method          \\ \midrule
LLM
                                & Self-refine~\cite{self-refine}, SelfCheck~\cite{selfcheck}, Reflexion~\cite{reflexion}, REFINER~\cite{refiner} \\ \midrule
Environment   
                                & SayPlan~\cite{sayplan}, Re-prompt~\cite{re-prompting}, Inner Monologue~\cite{inner-monologue},  GITM~\cite{gitm}, DEPS~\cite{deps}          \\ \midrule
VLM 
                                & DEPS ~\cite{deps}, LLM-Planner~\cite{llm-planner}, Inner Monologue~\cite{inner-monologue} \\ \midrule
Interpreter               & Self-debug~\cite{self-debug},                                       INTERVENOR~\cite{wang2023intervenor}, CRITIC~\cite{critic}, MAF~\cite{maf} \\ \midrule
Heuristic Rule                  & Reflexion~\cite{reflexion}      \\\midrule
Human          & Inner Monologue~\cite{inner-monologue}, DEPS~\cite{deps}          \\ \midrule
Search Engine & CRITIC~\cite{critic}, Verify-and-Edit~\cite{verifyandedit} \\ \bottomrule
\end{tabular}
\caption{Feedback source used by existing works.}
\label{tab:feedback}
\end{table}

In CoT, the results generated by LLMs are assumed to be correct hence there is no evaluation process for the sub-goal results. 
The negative impact of hallucination becomes more severe for sub-goal results because errors may propagate along the pipeline and result in huge deviations from the correct answer. Thus, it is necessary to get feedback for sub-goal results at each step and correct them in time through resampling. The types of feedback sources are summarized in Table~\ref{tab:feedback}.  Note that the methods we review in this section can be applied to evaluate the overall goal achievement as well. 

\begin{figure*}[t]
\centering
\includegraphics[width=0.98\textwidth]{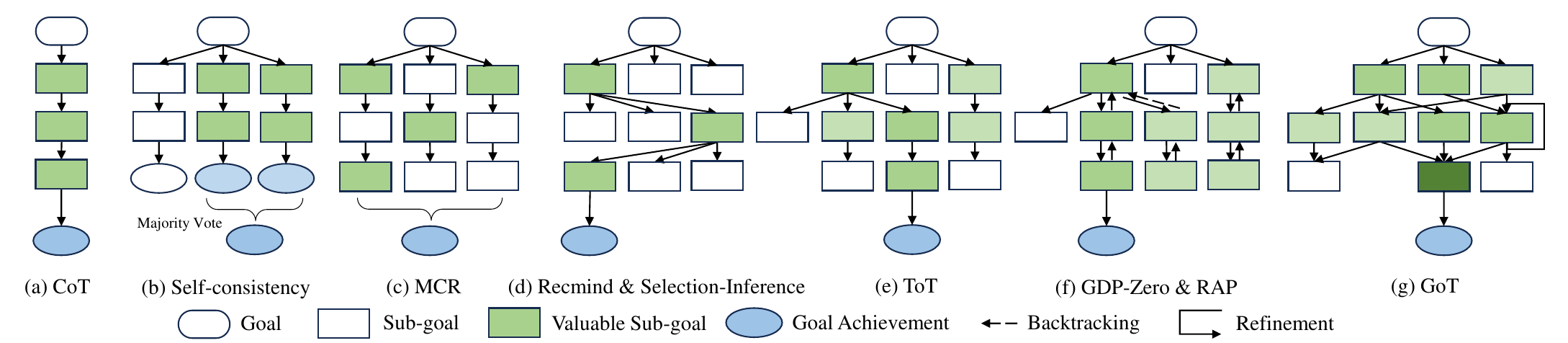}
\caption{Illustration of approaches for valuable sub-goal selection. (a) CoT selects all sub-goals in one sub-goal sequence. (b) Self-consistency, a variant of CoT, selects sub-goals based on majority votes. (c) MCR, a variant of CoT, selects sub-goals from multiple sub-goal sequences. (d) Recmind and Selection-Inference, variants of CoT, select one sub-goal from candidates at each step. (e) ToT explores sub-goals from a tree structure space. (f) GDP-Zero and RAP, variants of ToT, introduce backpropagation to ToT to balance exploration and exploitation. (g) GoT models all sub-goals in a graph structure space.}
\label{fig:sub-goal selection}
\vspace{-7pt}
\end{figure*}

\paragraph{Feedback from LLMs.}
Self-refine~\cite{self-refine, self-criticism} is the first work addressing the evaluation of sub-goal results. After getting the output of LLMs, Self-refine feeds the output combined with task-specific prompts to the same LLM to get feedback for the output. 
Then, the feedback and initial output are fed together to LLMs for output refinement. 
The refinement process iterates with all of the past feedback and refinement appended to the prompt. 
SelfCheck~\cite{selfcheck} decomposes the Self-refine evaluation into finer-grained processes. First, actions and sub-goal results are put to LLMs to summarize their intention. Then, the goal combined with all previous sub-goals, selected actions, and sub-goal results serves as another input to let LLMs extract useful information from them. Finally, only being provided with useful information and summarized intention, LLMs generate the sub-goal result again. This result will be compared with the original result. 
If the regeneration result supports the original sub-goal result, the original result is considered correct, and vice versa.
In Reflexion~\cite{reflexion} and REFINER~\cite{refiner}, the authors try using another instantiation of an LLM to obtain feedback for the sub-goal result.

\paragraph{Feedback from external tools.}
In contrast to self-evaluation, we could seek the help of external evaluators to provide feedback.

Some methods~\cite{self-debug,wang2023intervenor,sayplan,re-prompting} leverage the error messages that are pre-built into either the programming languages or the program of virtual environments to evaluate sub-goal results.
Self-debug~\cite{self-debug}, MAF~\cite{maf} and INTERVENOR~\cite{wang2023intervenor} leverages unit tests and executors to gain feedback on generated code snippets. 
The authors also empirically show that leveraging unit test error messages leads to superior performance than LLM's self-reflection.
This finding indicates that external evaluators may be more beneficial than self-reflection thanks to the more precise feedback~\cite{critic}. 
SayPlan~\cite{sayplan} and Re-prompting~\cite{re-prompting} directly leverage the signs of success or failure with error messages from the environment as feedback to revise their actions. GITM~\cite{gitm} designs a structured feedback template to get detailed knowledge of the current situation and all of the information can be obtained from environment API returns.

However, sometimes it is hard to define error messages in advance. To further address it, LLM-Planner~\cite{llm-planner} explores the possibility of using Vision-Language Models (VLMs) to provide feedback by describing the environment. Inner Monologue~\cite{inner-monologue} and DEPS~\cite{deps} are combinations of SayPlan and LLM-Planner, 
where VLMs additionally perform visual question answering to enable a finer-grained human-computer interaction for valuable information. 
Reflexion~\cite{reflexion} proposes various heuristic functions to score different tasks (e.g. exact match for reasoning). 
CRITIC~\cite{critic} and Verify-and-Edit~\cite{verifyandedit} employ search engines to obtain true information as feedback from the web.

\subsection{Further Selection of Valuable Sub-goals}
 
The generated sub-goals in stage (1) may be wrong or irrelevant to overall goal achievement.
To address this problem, we could ask LLMs to explore several sub-goals in each step or even several sub-goal sequences and then only select those valuable candidates. In this section, we introduce methods for such sub-goal post-processing, which can be divided into three stages of development, CoT and its variants, Tree of thoughts (ToT)~\cite{tot} and its variants, and finally Graph of thoughts (GoT)~\cite{got}. An illustration of these approaches is shown in Fig.~\ref{fig:sub-goal selection} to better distinguish these methods. 

\paragraph{Chain-of-thought and its variants.}

We consider CoT as simply selecting all sub-goals since LLMs have access to all of the sub-goals during the generation process, which makes it prone to irrelevant sub-goals or wrong sub-goal results.
Four works are proposed to improve the robustness of CoT. Self-consistency~\cite{self-consistency} prompts LLMs several times via CoT and then conducts majority votes based on the similarity of different outputs from each COT sequence to decide valuable sub-goal sequences. The result from the chosen sequence is considered the final result. MCR~\cite{mcr} improves Self-consistency by selecting and combining sub-goals from different sequences. After getting several sub-goal sequences, it uses LLMs to extract valuable sub-goals to form a new sub-goal sequence, dubbed meta-reasoning path, and predict the result based on the meta path. Recmind~\cite{recmind} and Selection-Inference~\cite{select-inference} allow multiple sub-goal candidates at each decomposition step and choose the most valuable one based on LLM self-evaluation. 

\paragraph{Tree of thoughts and its variants.} 
ToT~\cite{tot} advances over CoT by enabling the exploration and comparison of sub-goals based on a tree structure.
To make global choices, at each sub-goal when traversing the tree, a decision can be made that involves backtracking or looking ahead based on the implementation of traditional searching algorithms like breadth-first search. For the comparison of sub-goals, ToT solely relies on the self-evaluation of LLMs. CoT and its variants can be considered special cases of ToT, where each node only has one child node. 

RAP~\cite{rap} and GDP-Zero~\cite{gdp-zero} improve from ToT's heuristic-based search to Monte Carlo Tree Search (MCTS). Compared with ToT, MCTS has an additional back-propagation step, where the score value that indicates the goal achievement is back-propagated to update the value score of all sub-goals in the sub-goal sequence. This operation strikes a proper balance between exploration and exploitation to find valuable sub-goals efficiently, which makes MCTS outperform heuristic-based search algorithms on complex or less structured tasks.

\paragraph{Graph of thoughts.}
GoT~\cite{got} advances ToT by modeling all of the LLM-generated sub-goals as a graph. While in ToT, only sub-goals in a sequence or at the same step can interact with each other through comparison, lookahead, or backtracking, GoT allows the synergy of arbitrary sub-goals. What's more, GoT supports sub-goal aggregation to generate a new sub-goal and sub-goal refinement based on other dependent sub-goals.
Finally, valuable sub-goals are selected via LLM self-evaluation on the constructed graph.

\section{Discussion}

\paragraph{Application.} 

After understanding how each method leverages goal-oriented prompt engineering to achieve goals with LLMs, we find it insightful to provide a bird-eye view of the surveyed paper. In Table~\ref{tab:app}, we classify these papers by their targeted tasks and the stages involved in our goal-oriented framework, to help readers better understand the current progress\footnote{Note that here we only introduce representative tasks. For more comprehensive information, we maintain a list at \url{https://github.com/Alex-HaochenLi/Goal-oriented-Prompt-Engineering}.}.

The 50 representative works in our survey cover 11 tasks, including arithmetic reasoning (18 works), commonsense reasoning (5 works), symbolic reasoning (3 works), logical reasoning (5 works), planning in virtual/real environment (12 works), multihop question answering (8 works), open-domain question answering (1 work), code generation (4 works), dialogue systems (6 works), and recommender systems (3 works). We could see that 46\% of the works focus on reasoning, 24\% of the works are for planning and the remaining 30\% of the works are for other tasks.

For reasoning tasks, all the surveyed works involve only a single stage, with goal decomposition, action execution, sub-goal result evaluation, and valuable sub-goal selection holding 27\% equally. There is potential for exploring the synergy of stages, which we will discuss in Section~\ref{sec:future}.
For planning tasks, goal decomposition and action selection share 27\%, and sub-goal result evaluation shares the rest 46\%. 
No surveyed work for planning tasks involves the action execution stage and valuable sub-goal selection stage. 
This is because the currently available evaluation benchmarks are based on either simple real environments or carefully designed virtual environments. Actions are either easy or can be assured to be perfectly executed. Nonetheless, as the field develops, we can expect more complicated environments where there should be attention paid to action execution and valuable sub-goal selection stage.

For those minority tasks like code generation and recommendation, most of the stages remain unexplored. With the goal-oriented framework proven effective in reasoning and planning tasks, it should be valuable to explore the potential of the framework on other tasks and we will discuss it in Sec~\ref{sec:future}.

\paragraph{Performance.}

Furthermore, We compare the performance of the surveyed methods in Appendix~\ref{app:performance comparison} (Table~\ref{tab:arith}, \ref{tab:commonsense}, \ref{tab:sym}, \ref{tab:logic}, \ref{tab:plan}, \ref{tab:multihop}, \ref{tab:code}, \ref{tab:rec}), and there are some noteworthy findings. 
In arithmetic reasoning, compared with standard I-O prompting, on average goal decomposition (CoT) improves by 22.6\%, sub-goal evaluation (Self-refine~\cite{self-refine}) improves by 21.1\%, and valuable sub-goal selection (Self-consistency~\cite{self-consistency}) improves by 32.5\%. While valuable sub-goal selection improves the most in arithmetic reasoning, the improvement brought by goal decomposition outperforms sub-goal selection in symbolic reasoning, which indicates that the variations of improvement have relations to task characteristics.
Besides, we observe a gradual increase in performance when integrating additional stages into the method. PoT~\cite{pot} can be considered as a combination of CoT and action execution, it further improves CoT by 14.7\% in arithmetic reasoning. Similarly, Sayplan~\cite{sayplan} in planning tasks, equipped with goal decomposition, action selection, and sub-goal evaluation, outperforms solely sub-goal evaluation (LLM-Planner~\cite{llm-planner}) by 73.3\% in excitability. Thus, we could expect further improvement from stage synergy.

During data collection, we find it hard to compare between methods due to the variety of prompt templates, the number of in-context examples included, and the decoding strategies, even if the methods are applied to the same LLM in the same zero-/few-shot setting. With our survey, we hope to 
promote a unified evaluation procedure for each task for fair comparison of future methods.

\section{Challenges \& Opportunities}
\label{sec:future}
In this section, we discuss the challenges and opportunities of goal-oriented prompt engineering based on our findings. 

\paragraph{Hierarchical Goal Decomposition.}
Existing works only decompose the goal into one layer of sub-goals, ignoring the fact that the sub-goal itself can still be complex and multifaceted. 
To make sure each sub-goal is simple enough to be handled by LLMs, researchers are now exploring hierarchical decomposition, which further decomposes sub-goals into simpler "sub-sub-goals". \cite{li2023take} is a pioneering work that constructs a dataset for the script generation task focusing on hierarchical decomposition. 
GITM~\cite{gitm} employs the idea of hierarchical decomposition by utilizing a pre-defined sub-goal tree to connect a sub-goal with even simpler ones. 
Though a significant success rate increase by employing hierarchical decomposition is reported, a pre-defined tree is not always accessible. Besides, relying solely on one tree may not be able to handle more complex problems, which leaves room for the introduction of graphs for further improvement.

\paragraph{Efficiency.}

The current design of goal-oriented prompt engineering focuses on improving the accuracy of models. Nevertheless, efficiency also plays a crucial role in the real-world application of the methods. Researchers should consider the trade-off between accuracy and efficiency when necessary.

We propose to consider two perspectives of efficiency measurement, task-agnostic efficiency and task-specific efficiency. Task-agnostic efficiency aims at measuring the efficiency of model targets for any task, for instance, the total number of prompting times before reaching the sub-goal or final goal.
On the other hand, task-specific efficiency is designed to measure efficiency that appears only in certain tasks. 
As an example, the number of iterations that a certain method prompts LLMs would affect the speed of plan generation in planning tasks. Such task-agnostic efficiency is important in scenarios like autonomous driving path planning when we need a quick response for better user experiences. On the other hand, the output plan needs to fulfill other criteria, such as minimizing traveling time or distance. Here, the time and distance are task-specific efficiency.

\paragraph{Stage Synergy.}

Overall there is a lack of systematic synergy of stages in any task.
We found two works that combine their methods, on one stage, with a method, on another stage, and show performance improvement~\cite{sayplan, pot}.
This suggests the power of synergizing the stages in the goal-oriented framework and careful integration of stages would have the potential for further improvement.

We take the arithmetic reasoning task as an example. To solve a math problem, a naive solution would be to follow Plan-and-solve~\cite{plan-and-solve} to decompose the math problem into a few logical steps. Then, following MWP~\cite{math-word-problem}, an operation prediction module can be applied to determine the correct mathematical operations. LLMs can select the operands for performing the operation. Then, LLMs could call an external calculator to calculate the equation to ensure an accurate result. With the result, we could follow Reflexion~\cite{reflexion} to ask LLMs to double-check the intermediate steps again. 
We could further repeat the described process to get multiple sequences of sub-goals and follow MCR~\cite{mcr} to remove less valuable equations for more accurate answers. 
Of course, a naive combination of methods from all stages is heavy, requiring additional attention paid to efficiency. 
Additionally, it is also important to consider the between-stages interaction.
For the same math problem, if we apply a calculator for calculation (action execution) to guarantee a correct answer, we should be careful when designing the result evaluation stage. If we use LLMs for result evaluation, the correct result from the calculator may inversely be considered as wrong due to the hallucination problem.

\paragraph{Broader Applications.}

From our summarization, we observe an absence of study in certain stages for some tasks, such as code generation and recommender system. There is potential to improve current methods by studying the unexplored stages. Moreover, goal-oriented prompt engineering has the potential for broader applications beyond the 11 tasks summarized in this paper.

In code generation, all of the surveyed methods focus on leveraging sub-goal evaluation to refine the code generation result.
Additional refinement may be done by incorporating other stages. Given a description of the desired code snippets, we could first decompose it into several required sub-functions, and then select appropriate algorithms to implement them. As for the implementation stage of these sub-functions, we could search relevant documents of potential well-written function APIs for reference instead of generating the code fully based on LLMs' knowledge. APIs in external libraries are regularly updated which can perform better than the static knowledge stored in LLMs.

In recommender systems, works have been proposed for enhancing action execution and sub-goal result evaluation. There is an absence of work on action selection.
Take keeping user retention as the overall goal, one sub-goal might be to make sure that users are recommended with their preferred items.
There are various actions we could take to predict the preference of a user for a certain item, such as inferring based on the click rate of similar items and inferring based on the watching time of relevant advertisements. Careful design on dynamically selecting between available actions based on the characteristics of the user may enable more accurate prediction.

For tasks beyond the scope of this paper, here we take code review as an example. Code review is a crucial but cumbersome task in software development. It ensures the quality and maintainability of code but requires a reviewer to spend a lot of time doing the systematic examination. Automatic code review can help save such significant effort. Due to the coding ability shown by recent LLMs, there is potential to surpass existing static analysis tools in automatic code review if equipped with our goal-oriented framework.
We found one primary study~\cite{codereview}, which focuses on action selection, already shows that reducing the action space by identifying the type of request changes can significantly improve the success rate of ChatGPT. 
Here we give another possibility which covers more stages of our framework.
With reviewing code as the goal, sub-goals can be documentation quality, function reusability, code style consistency, variable naming readability, and so on. 
Taking variable naming readability as the sub-goal, we can have LLMs select action between checking the self-consistency of the variable names or comparing those names with the pre-defined naming convention (the action selection stage).
Subsequently, LLMs can execute the action by either calling an external tool for similarity measurement or following given heuristic rules (the action execution stage). 
After going through all the sub-goals (i.e. reviewing criteria), LLMs can revise the given code based on reviews from each perspective.

\section{Conclusion}
In this paper, we provide a review of existing research on LLMs prompting from a goal-oriented perspective. We explicitly summarize the task-agnostic internal steps involved in designing and refining prompting procedures into five stages, classify the surveyed methods by their targeted tasks, and provide a comprehensive analysis on task coverage and performance. 
Though significant progress has been made in this field, further research is needed to fully unleash the potential by studying hierarchical goal decomposition, the synergy of stages, application to other tasks, and considering efficiency problems.
We hope this paper can serve as an overview of the current state of goal-oriented prompt engineering and stimulate further research on this important topic.

\bibliography{custom}

\appendix

\section{Related Surveys}
\label{relatedsurvey}
While several surveys have explored LLM prompting, none have examined it from a goal-oriented perspective or explicitly summarized the task-agnostic internal steps involved in designing and refining prompting procedures.
Some surveys summarize studies that focus on specific tasks or problems. \cite{reasoningsurvey1,reasoningsurvey2, lu-etal-2023-survey, yang2023logical} focus solely on reasoning tasks and \cite{planningsurvey} on planning tasks. \cite{hallucinationsurvey} deals with the hallucination problem of LLMs.
Other studies focus on generally applicable prompting but different sub-fields.
\cite{simplesurvey1, simplesurvey2} simply listed out recent techniques without systematic categorization. \cite{yu2023towards, chu2023survey} study CoT and its XoT variants.
\cite{promptingframeworksurvey} focus on the prompting framework for facilitating the interaction between LLMs with external worlds.
\cite{oldsurvey} focuses on methods for learnable continuous prompt, while our work studies discrete prompt. 
As many of the will-functioned LLMs are not open-sourced (e.g. GPT-3.5, GPT4), discrete prompts became the focus of recent studies on LLM prompting and therefore we choose to focus on discrete prompting.

\section{Performance Comparison}
\label{app:performance comparison}

In this section, we compare the performance of surveyed methods under several tasks, including arithmetic reasoning (Table~\ref{tab:arith}), commonsense reasoning (Table~\ref{tab:commonsense}), symbolic reasoning (Table~\ref{tab:sym}), logical reasoning (Table~\ref{tab:logic}), planning (Table~\ref{tab:plan}), multi-hop question answering (Table~\ref{tab:multihop}), code generation (Table~\ref{tab:code}), and recommender systems (Table~\ref{tab:rec}). The empirical results are collected from corresponding papers or other papers where they serve as baselines. Here, the benchmarks we chose are the ones on which the majority of methods are evaluated. For a full list of benchmarks, we maintain it at \url{https://github.com/Alex-HaochenLi/Goal-oriented-Prompt-Engineering}. Note that due to variations in experimental settings like prompt template, decoding strategy, and numbers of in-context examples, even the methods that share the same zero-/few-shot setting and LLM may not be fairly comparable. Therefore, these tables only provide a rough trend of performance, and we don't make conclusions stating which method works best in certain contexts. We mainly aim to show that the introduction of goal-oriented prompt design can consistently bring improvement.

\begin{table*}[]
\centering
\resizebox{\textwidth}{!}{%
\begin{tabular}{@{}lccccccccc@{}}
\toprule
\multirow{2}{*}{Methods} & \multirow{2}{*}{Stage} & \multirow{2}{*}{Setting} & \multirow{2}{*}{Model} & GSM8K & SVAMP & AQuA & AddSub & MultiArith \\ 
 & & & & \cite{gsm8k} & \cite{svamp} & \cite{aqua} & \cite{addsub} & \cite{multiarith} \\
\midrule
Zero-shot CoT~\cite{zero-shot-cot} & \ding{172} & zero-shot & text-davinci-003                     & 56.4  & 69.9   & 38.9 & 85.3   & 83.8       \\
Few-shot CoT~\cite{cot} & \ding{172} & few-shot & text-davinci-003 & 58.4 & 80.3 & 48.4 & 91.6 & 93.6 \\
Plan-and-solve~\cite{plan-and-solve}  & \ding{172} & zero-shot & text-davinci-003        & 59.3  & 75.7    & 46.0 & 92.2   & 91.8       \\ 
Least-to-most~\cite{least-to-most} & \ding{172} & few-shot & code-davinci-002 & 38.3 & 80.3 & 40.6 & - & 74.0 \\
PoT~\cite{pot} &  \ding{174} & zero-shot   & text-davinci-003               & 57.0  & 70.8      & 43.9 & 85.1      & 92.2       \\
Toolformer~\cite{toolformer} & \ding{174} & zero-shot & GPT-J & - & 29.4 &- &- &- \\
PoT~\cite{pot} & \ding{174} & few-shot & code-davinci-002 & 71.6 & 85.2 & 54.1 & - & - \\
PAL~\cite{pal} & \ding{174} & few-shot & code-davinci-002 & 72.0 & 79.4 & - & 92.5 & 99.2 \\
Faithful CoT~\cite{faithfulcot} & \ding{174} & few-shot & code-davinci-002 & 72.3 & 83.4 & 47.2 & - & 98.8 \\
RAP~\cite{rap} & \ding{175} & few-shot & LLaMA-33B & 40.0 &- &- &- &- \\
Self-refine~\cite{self-refine} & \ding{175} & few-shot & text-davinci-003 & 64.1 & 67.6 & - & - & - \\
Self-consistency~\cite{self-consistency} & \ding{176} & few-shot & code-davinci-002 &         78.0 & 86.8 & 52.0 & 91.6 & 100.0      \\
\bottomrule
\end{tabular}
}
\caption{Performance comparison on arithmetic reasoning. The evaluation metric is solve rate (\%).}
\label{tab:arith}
\end{table*}

\begin{table*}[]
\centering
\resizebox{\textwidth}{!}{%
\begin{tabular}{@{}llccccc@{}}
\toprule
\multirow{2}{*}{Metric}    & \multirow{2}{*}{Methods} & \multirow{2}{*}{Stage} & \multirow{2}{*}{Setting} & \multirow{2}{*}{Model} &  CSQA      & StrategyQA      \\ 
 & & & & & \cite{csqa} & \cite{strategyqa} \\ \midrule
\multirow{3}{*}{Solve rate} & Zero-shot CoT~\cite{zero-shot-cot}        & \ding{172} & zero-shot & text-davinci-003              & 65.2      & 63.8            \\
& Plan-and-solve~\cite{plan-and-solve}  & \ding{172} & zero-shot  &   text-davinci-003      & 71.9      & 65.4            \\ 
& Few-shot CoT~\cite{cot}        & \ding{172} & few-shot & text-davinci-003              & 78.3      & 71.2            \\
& Few-shot CoT~\cite{cot}        & \ding{172} & few-shot & code-davinci-002              & 79.0      & 73.4            \\  
                            & Self-consistency~\cite{self-consistency} & \ding{176} & few-shot & code-davinci-002       & 81.5      & 79.8            \\\midrule
\multirow{3}{*}{F1}   & CoT~\cite{cot} & \ding{172} & few-shot & code-davinci-002 & - & 70.0 \\ 
& Self-ask~\cite{self-ask} & \ding{172} & few-shot & code-davinci-002                & -         & 69.3            \\ 
& MCR~\cite{mcr} & \ding{176} & few-shot & code-davinci-002 &                      -         & 73.6            \\
                            & Self-consistency~\cite{self-consistency}  & \ding{176} & few-shot & code-davinci-002        & -         & 72.2            \\ \bottomrule              
\end{tabular}
}
\caption{Performance comparison on commonsense reasoning under solve rate (\%) and F1 (\%). F1 is computed by treating prediction and ground truth answers as bags of tokens and computing their precision and recall.}
\label{tab:commonsense}
\end{table*}

\begin{table*}[]
\centering
\small
\begin{tabular}{@{}lccccc@{}}
\toprule
\multirow{2}{*}{Methods}  & \multirow{2}{*}{Stage} & \multirow{2}{*}{Setting} & \multirow{2}{*}{Model} & last letter concatenation  & coinflip          \\ 
 & & & & \cite{cot} & \cite{cot} \\ \midrule
I-O Prompting & - & few-shot & text-davinci-002 &  5.8 & 49.0 \\
Zero-shot CoT~\cite{zero-shot-cot}   & \ding{172}     & zero-shot & text-davinci-003              & 64.8                      & 96.8                 \\
Plan-and-solve~\cite{plan-and-solve}    & \ding{172} & zero-shot & text-davinci-003        & 75.2                       & 99.6                 \\
Few-shot CoT~\cite{cot}   & \ding{172}     & few-shot & text-davinci-003              & 70.6                       & 100.0                 \\
Few-shot CoT~\cite{cot} & \ding{172} & few-shot & code-davinci-002         & 70.4                      & 99.0                  \\
Least-to-most~\cite{least-to-most} & \ding{172} & few-shot & code-davinci-002 & 94.0 & - \\
Self-consistency~\cite{self-consistency} & \ding{176} & few-shot & code-davinci-002         & 73.4                       & 99.5                  \\
 \bottomrule
\end{tabular}
\caption{Performance comparison on symbolic reasoning. The evaluation metric is solve rate (\%).}
\label{tab:sym}
\end{table*}

\begin{table*}[]
\centering
\small
\begin{tabular}{@{}lccccc@{}}
\toprule
\multirow{2}{*}{Methods}     & \multirow{2}{*}{Stage} & \multirow{2}{*}{Setting} & \multirow{2}{*}{Model} & FOLIO & ProofWriter \\ 
 & & & & \cite{folio} & \cite{proofwriter} \\ \midrule
I-O Prompting & - & zero-shot  &  gpt-3.5-turbo     & 48.4    & 36.4                   \\            
CoT~\cite{cot} & \ding{172} & few-shot  &  gpt-3.5-turbo     & 54.9      & 43.6                    \\
Logic-LM~\cite{logic-lm} & \ding{174} & few-shot  & gpt-3.5-turbo & 62.7     & 58.3                     \\
LINC~\cite{linc} & \ding{174} & few-shot & gpt-3.5-turbo  & 62.6       & 96.4                     \\
LINC~\cite{linc} & \ding{174} & few-shot & GPT-4  & 72.5       & 98.3                    \\
 \bottomrule
\end{tabular}
\caption{Performance comparison on logical reasoning. The evaluation metric is solve rate (\%).}
\label{tab:logic}
\end{table*}

\begin{table*}[]
\centering
\resizebox{\textwidth}{!}{%
\begin{tabular}{@{}llccccc@{}}
\toprule
\multirow{2}{*}{Dataset}     & \multirow{2}{*}{Methods} & \multirow{2}{*}{Stage} & \multirow{2}{*}{Setting} & \multirow{2}{*}{Model} & \multicolumn{2}{c}{Evaluation Metric} \\ \cmidrule(l){6-7} 
                             &                          &                        &                          &                        & Executability      & Correctness      \\ \midrule
\multirow{2}{*}{Home+Office} & LLM+P~\cite{llm+p}                    & \ding{172}                      & few-shot                 & GPT-4                  & 0.0                  & 33.3             \\
                             & LLM-Planner~\cite{llm-planner}              & \ding{175}                      & few-shot                 & GPT-4                  & 13.3               & 66.7             \\
\cite{sayplan}                             & SayPlan~\cite{sayplan}                  & \ding{172}\ding{173}\ding{175}                    & few-shot                 & GPT-4                  & 86.6               & 73.3             \\ \midrule
\multirow{2}{*}{VirtualHome} & Zero-shot Planner~\cite{zero-shot-planner}        & \ding{173}                      & few-shot                 & GPT2-large             & 16.4               & 33.3             \\
                             & Zero-shot Planner~\cite{zero-shot-planner}        & \ding{173}                       & few-shot                 & Codex                  & 78.4               & 46.1             \\
\multirow{2}{*}{\cite{virtualhome}}                             & SALP~\cite{generating-executable-action}                     & \ding{173}                       & few-shot                 & GPT2-large             & 50.8               & 49.0             \\
                             & Re-prompting~\cite{re-prompting}             & \ding{173}                       & few-shot                 & Codex                  & 98.9               & 41.8             \\ \bottomrule
\end{tabular}
}
\caption{Performance comparison on planning in virtual environment. The evaluation metrics are executability (\%) and correctness (\%). Executability measures if the generated actions can be correctly parsed in the environment, while Correctness is a human-annotated metric that assesses whether the given steps could accomplish the given task.}
\label{tab:plan}
\end{table*}

\begin{table*}[]
\centering
\resizebox{\textwidth}{!}{%
\begin{tabular}{@{}llcccccc@{}}
\toprule
\multirow{2}{*}{Metric}   & \multirow{2}{*}{Methods} & \multirow{2}{*}{Stage} & \multirow{2}{*}{Setting} & \multirow{2}{*}{Model} &  2WikiMultihopQA & Bamboogle & HotpotQA  \\ 
 & & & & & \cite{2wikimultihopqa} & \cite{bamboogle} & \cite{hotpotqa} \\ \midrule
\multirow{3}{*}{Acc.} & 
I-O Prompting & - & few-shot & Codex & 25.4 & 17.6 & - \\
 & CoT~\cite{cot} & \ding{172} & few-shot &        Codex              & 29.8            & 46.4    & -        \\
                          & Least-to-most~\cite{least-to-most}  & \ding{172} & few-shot  &  Codex         & 29.0            & -    & -        \\
                          & Self-ask~\cite{self-ask} & \ding{172} & few-shot  &   Codex              & 40.1            & 60.0    & -        \\ \midrule
\multirow{4}{*}{F1}       & CoT~\cite{cot} & \ding{172} & few-shot  & Codex & 67.2 & 64.7 & 56.4 \\
                          & Self-ask~\cite{self-ask}  & \ding{172} &few-shot  &   Codex             & 63.8            & 64.6       & 50.2     \\
                          & DecomP~\cite{decomposed-prompting} &\ding{172} & zero-shot &   Codex                & 64.1            & 25.4    & 49.9     \\ 
                          & MCR~\cite{mcr}  & \ding{176} & few-shot &  Codex                   & 67.9            & 66.5       & 57.0     \\
                          & Self-consistency~\cite{self-consistency} & \ding{176} &few-shot  &  Codex       & 65.4            & 65.0       & 56.4     \\
                          \bottomrule 
\end{tabular}
}
\caption{Performance comparison on multihop question answering under accuracy (\%) and F1 (\%). F1 is computed by treating prediction and ground truth answers as bags of tokens and computing their precision and recall.}
\label{tab:multihop}
\end{table*}

\begin{table*}[]
\centering
\small
\begin{tabular}{@{}lccccc@{}}
\toprule
\multirow{2}{*}{Methods}     &\multirow{2}{*}{Stage} & \multirow{2}{*}{Setting} & \multirow{2}{*}{Model} & HumanEval & MBPP \\
 & & & & \cite{humaneval} & \cite{mbpp} \\ \midrule
Zero-shot I-O Prompting & - & zero-shot  &  gpt-3.5-turbo     & 62.2      & 41.6                    \\
Few-shot I-O Prompting & - & few-shot  &  gpt-3.5-turbo     & 65.2      & 40.6                     \\
CoT~\cite{cot} & \ding{172} & zero-shot  &  gpt-3.5-turbo     & 66.5      & 48.8                     \\
Self-refine~\cite{self-refine} & \ding{175} & zero-shot  & gpt-3.5-turbo & 65.2      & 48.8                     \\
INTERVENOR~\cite{wang2023intervenor} & \ding{175} & zero-shot & gpt-3.5-turbo & 75.6      & 69.8                     \\
Reflexion~\cite{reflexion} & \ding{175} & zero-shot & GPT-4  & 91.0        & 77.1                     \\
Self-debug~\cite{self-debug} & \ding{175} & few-shot & code-davinci-002 & -         & 75.6                     \\
 \bottomrule
\end{tabular}
\caption{Performance comparison on code generation. The evaluation metric is Pass@1 (\%).}
\label{tab:code}
\end{table*}

\begin{table*}[]
\centering
\small
\begin{tabular}{@{}lccccc@{}}
\toprule
\multirow{2}{*}{Methods}             & \multirow{2}{*}{Stage} & \multirow{2}{*}{Setting} & \multirow{2}{*}{Model} & Amazon Review & Yelp \\
 & & & & \cite{amazonreview} & \cite{yelp} \\ \midrule
I-O Prompting~\cite{liu2023chatgpt} & - & zero-shot & gpt-3.5-turbo & 1.1897 & 1.2359 \\
CoT~\cite{cot}  & \ding{172}               & zero-shot & gpt-3.5-turbo & 0.8612                 & 1.1673                   \\
Selection-Inference~\cite{select-inference} & \ding{176} & zero-shot & gpt-3.5-turbo & 0.7883                 & 1.0009                  \\ \midrule
I-O Prompting~\cite{liu2023chatgpt} & - & few-shot & gpt-3.5-turbo & 0.7327 & 1.0016 \\
CoT~\cite{cot}  & \ding{172}               & few-shot & gpt-3.5-turbo & 0.7167                 & 0.9794                   \\
ToT~\cite{tot}  & \ding{176}               & few-shot & gpt-3.5-turbo & 0.7059                 & 0.9766                   \\
Selection-Inference~\cite{select-inference} & \ding{176} & few-shot & gpt-3.5-turbo & 0.6892                 & 0.9698                   \\ \bottomrule
\end{tabular}
\caption{Performance comparison on rating prediction in recommender system. The evaluation metric is Mean Absolute Error. Here I-O prompting means that LLMs are directly prompted for the final prediction.}
\label{tab:rec}
\end{table*}

\begin{table*}[]
\centering
\small
\begin{tabular}{@{}lllp{9.0cm}@{}}
\toprule
Application                         & Sub-category                           & Stage                & Method                                                                                                                                                                           \\ \midrule
\multirow{13}{*}[-7em]{Reasoning}         & \multirow{5}{*}[-3.5em]{Arithmetic}  & \multirow{1}{*}[-1.2em]{\ding{172}}           & CoT~\cite{cot}, Least-to-most~\cite{least-to-most}, Plan-and-solve~\cite{plan-and-solve}, Successive Prompt~\cite{successive-prompt}, RoT~\cite{rot}                                             \\ \cmidrule(l){3-4} 
                                    &                                        & \ding{173}           & MWP~\cite{math-word-problem}                                                                                                                                                     \\ \cmidrule(l){3-4} 
                                    &                                        & \multirow{1}{*}[-1.2em]{\ding{174}}           & PoT~\cite{pot}, Toolformer~\cite{toolformer}, Faithful CoT~\cite{faithfulcot}, PAL~\cite{pal}, MathPrompter~\cite{mathprompter}                                                                                                                                     \\ \cmidrule(l){3-4} 
                                    &                                        & \multirow{1}{*}[-1.2em]{\ding{175}}           & Self-refine~\cite{self-refine}, RAP~\cite{rap}, SelfCheck~\cite{selfcheck}, REFINER~\cite{refiner}, CRITIC~\cite{critic}, MAF~\cite{maf}                                                                                                       \\ \cmidrule(l){3-4} 
                                    &                                        & \ding{176}           & Self-consistency~\cite{self-consistency}                                                                                                                                         \\ \cmidrule(l){2-4} 
                                    & \multirow{3}{*}[-1.2em]{Commonsense} & \ding{172}           & CoT~\cite{cot}, Plan-and-solve~\cite{plan-and-solve}                                                                                                                             \\ \cmidrule(l){3-4} 
                                    &                                        & \ding{174}           & Toolformer~\cite{toolformer}                                                                                                                                                     \\ \cmidrule(l){3-4} 
                                    &                                        & \ding{176}          & Self-consistency~\cite{self-consistency}, MCR~\cite{mcr}                                                                                                                         \\ \cmidrule(l){2-4} 
                                    & \multirow{3}{*}[-0.7em]{Symbolic}    & \ding{172}           & Plan-and-solve~\cite{plan-and-solve}                                                        \\ \cmidrule(l){3-4} 
                                    &                                        & \ding{174}           & PAL~\cite{pal}                                                                                    \\ \cmidrule(l){3-4} 
                                    &                                        & \ding{176}           & Self-consistency~\cite{self-consistency}                                                                                                                                         \\ \cmidrule(l){2-4} 
                                    & \multirow{4}{*}[-0.7em]{Logical}     & \ding{173}           & DecomP~\cite{decomposed-prompting}                                                    \\ \cmidrule(l){3-4} 
                                    &                                        & \ding{174}           & LINC~\cite{linc}, Logic-LM~\cite{logic-lm}                                                                                                                                                       \\ \cmidrule(l){3-4} 
                                    &                                        & \ding{175}           & RAP~\cite{rap}                                                                                                                                                                   \\ \cmidrule(l){3-4} 
                                    &                                        & \ding{176}           & Selection-Inference~\cite{select-inference}                                                                                                                                      \\ \midrule
\multirow{4}{*}[-4em]{Planning}           & \multirow{3}{*}[-3.5em]{Virtual Env}   & \multirow{1}{*}[-0.5em]{\ding{172}}           & LLM+P~\cite{llm+p}, SayPlan~\cite{sayplan}, DEPS~\cite{deps}, GITM~\cite{gitm}                                                                                                   \\ \cmidrule(l){3-4} 
                                    &                                        & \multirow{1}{*}[-1em]{\ding{173}}           & Zero-shot Planner~\cite{zero-shot-planner}, Re-prompting~\cite{re-prompting}, SALP~\cite{generating-executable-action}, SayPlan~\cite{sayplan},                                  \\ \cmidrule(l){3-4} 
                                    &                                        & \multirow{1}{*}[-1.7em]{\ding{175}}           & Re-prompting~\cite{re-prompting}, RAP~\cite{rap}, Reflexion~\cite{reflexion}, SayPlan~\cite{sayplan}, DEPS~\cite{deps}, GITM~\cite{gitm}, Inner Monologue~\cite{inner-monologue}, LLM-Planner~\cite{llm-planner} \\ \cmidrule(l){2-4} 
                                    & Real Env                     & \ding{175}           & Inner Monologue~\cite{inner-monologue}                                                                                                                                           \\ \midrule
\multirow{6}{*}[-1.7em]{Question Answering} & \multirow{5}{*}[-1.5em]{Multihop QA}           & \ding{172}           & DecomP~\cite{decomposed-prompting}, Self-ask~\cite{self-ask}                                                                                                                     \\ \cmidrule(l){3-4} 
                                    &                                        & \ding{173}           & PEARL~\cite{pearl}                                                                                                                                                               \\ 
                                    \cmidrule(l){3-4} 
                                    &                                        & \ding{174}           & Faithful CoT~\cite{faithfulcot}                                                                                                                                                       \\
                                    \cmidrule(l){3-4} 
                                    &                                        & \multirow{1}{*}[-0.5em]{\ding{175}}           & Reflexion~\cite{reflexion}, CRITIC~\cite{critic}, Verify-and-Edit~\cite{verifyandedit}                                                                                                                                                       \\ \cmidrule(l){3-4} 
                                    &                                        & \ding{176}           & MCR~\cite{mcr}                                                                                                                                                                   \\ \cmidrule(l){2-4} 
                                    & Open domain QA                         & \ding{174}           & Toolformer~\cite{toolformer}                                                                                                                                                     \\ \midrule
\multirow{1}{*}[-0.5em]{Code Generation}                     &                                        & \multirow{1}{*}[-0.5em]{\ding{175}}           & Reflexion~\cite{reflexion}, Self-debug~\cite{self-debug}, Self-refine~\cite{self-refine}, INTERVENOR~\cite{wang2023intervenor}                                                   \\ \midrule
\multirow{3}{*}[-1.0em]{Dialogue}           & \multirow{3}{*}{}                      & \multirow{1}{*}[-0.5em]{\ding{173}}           & ProCoT~\cite{proactive-cot}, RLP~\cite{rlp}, Cue-CoT~\cite{cue-cot}, SAFARI~\cite{safari}                                                                                                                                                      \\ \cmidrule(l){3-4} 
                                    &                                        & \ding{175}           & Self-refine~\cite{self-refine}                                                                                                                                                   \\ \cmidrule(l){3-4} 
                                    &                                        & \ding{176}           & GDP-Zero~\cite{gdp-zero},                                                                                                                                                        \\ \midrule
\multirow{2}{*}[-1.0em]{Recommendation}                      &    \multirow{2}{*}{}                                     & \multirow{1}{*}[-0.5em]{\ding{174}} & Recmind~\cite{recmind}, DOKE~\cite{doke}, InteRecAgent~\cite{InteRecAgent}  \\ \cmidrule(l){3-4} 
 & & \ding{176} & Recmind~\cite{recmind} \\
\bottomrule
\end{tabular}
\caption{Applicable tasks from existing works. The stage column indicates the stages involved in the corresponding methods with regard to the goal-oriented framework in Fig.~\ref{overview}. Note that the methods in the table are not exclusively applicable to the tasks listed in the table to which they belong. We list them based on the evaluated tasks in their original papers.}
\label{tab:app}
\end{table*}

\end{document}